# *Meta-*Diversity Search in Complex Systems, A Recipe for Artificial Open-Endedness ?

Mayalen Etcheverry[a,b], Bert Chan, Clément Moulin-Frier[a], Pierre-Yves Oudeyer[a]

[a]FLOWERS Team, Inria, Bordeaux, France

[b] Poietis, Pessac, France

The idea of the Minecraft Open-Endedness Challenge is to try and think how would we artificially approach *"open-endedness"* in the Minecraft environment [Grbic et al., 2020], put in another words it asks the question: *Can we build an artificial system that would be able to generate endless surprises if ran "forever" in Minecraft?* While there is not a single path toward solving that grand challenge you've never heard of, this article presents what we believe to be some working ingredients for the endless generation of novel increasingly complex artifacts in Minecraft.

## Overview

Our framework for an open-ended system includes two components: a **complex system** used to recursively grow and complexify artifacts over time, and a discovery algorithm that leverages the concept of **meta-diversity** search.

Since complex systems have shown to enable the emergence of considerable complexity from set of simple rules, we believe them to be great candidates to generate all sort of artifacts in Minecraft. Yet, the space of possible artifacts that can be generated by these systems is often unknown, challenging to characterize and explore. Therefore automating the long-term discovery of novel and increasingly complex artifacts in these systems is an exciting research field. To approach these challenges, we formulate the problem of meta-diversity search where an artificial "discovery assistant" incrementally learns a diverse set of representations to characterize behaviors and searches to discover diverse patterns within each of them. A successful discovery assistant should continuously seek for novel sources of diversities while being able to quickly specialize the search toward a new unknown type of diversity [Etcheverry et al. 2020].

To implement those ideas in the Minecraft environment, we simulate an artificial "chemistry" system based on Lenia continuous cellular automaton for generating artifacts, as well as an artificial "discovery assistant" (called Holmes) for the artifact-discovery process. Holmes incrementally learns a hierarchy of modular representations to characterize divergent sources of diversity and uses a goal-based intrinsically-motivated exploration as the diversity search strategy.

In this blogpost, we start with the problem formulation for building an open-ended algorithm capable of endlessly generating Minecraft creations. Then, we present what we decided to include as desirata properties in our open-ended framework and then proposes a possible implementation of those ideas in the Minecraft environment. Finally, we provide the database of discoveries made by our algorithm.

## Problem Formulation

Minecraft game environnement $E$ is made of $1m^3$ building blocks that players can use to construct any structure they desire (also called *artifacts*) with almost endless possibilities to express their creativity. The blocks exist in a spatio-temporal world which is 3-dimensional in space and where time advances with each gametick. There is a finite number of possible blocks (M=255) which are described by their *type* $m$ (eg: water, wood, air, etc.) and optionally by a *state* $s$ which further describes the block appearance (eg: texture, orientation) or behavior (eg: age, on/off state, counters).

Let's define the *spaces* of our problem:

- **State space** $\Sigma$: each "cell" of the world (x,y,z) can take a certain state $s^t$. The *visible* state in the Evocraft API world is limited to the block type $j \in \{1..M\}$. The *hidden* state is free of choice, for instance a feature vector $h \in \mathbb{R}^d$ can be attached to each "cell" of the world and can describe concrete physical attributes of the cell at this point in space in time (eg: mass or electric currant) as well as more abstract computational quantities.
- **Artifact space** $A$: an *artifact* $a$ in Minecraft (also called *phenotype*) is represented by its discrete time serie $a = \{a^0, \ldots, a^t, \ldots\}$ ($a^t = a^0$ for static artifacts) and $a^t$ is a set $\{(m, x, y, z, s)\}$ of elemental blocks where $m$ is the block type; $(x, y, z)$ is the spatial position; and $s$ is the block state.
- **Search space** $\Theta$: each artifact is the result of some sort of *design* or *generative* process where a *genome* $\theta$ determines the artifact initial state $a^0$ as well as the artifact development over time $(a^0, \ldots, a^t) \to a^{t+1}$. The choice of the genome $\theta$ and of the generative process $G : \Theta \to A$ (also called *genotype-to-phenotype* mapping) will define the search space $\Theta$ and the reachable artifact space $A$ of our system.
- **Solution space** $W$: the solution space is set of all possible *artifacts* $\{a \in E\}$ which can be generated by $G$.

Following that formalism, our open-ended framework is composed of two main components:

- an **artifact-generator** $G : \Theta \to A$ which should allow for artifacts with unbounded phenotypic complexity.
- an **artifact-discovery algorithm** $D$ that evolves a set of genomes and their corresponding artifacts $\{(\theta_1, a_1 = G(\theta_1)), \ldots, (\theta_N, a_N = G(\theta_N))\} \in W$ ($N \to \infty$ in the open-ended setting) and that should be able to continuously evolve novel and increasingly complex artifacts.

How to artificially construct the genome $\theta$, the genotype-to-phenotype mapping $G$ and to how to evolve sets of artifacts $\{(\theta, a)\} \in W$ complying with open-ended problems remains an open question to date. However, interdisciplinary research at the crossroad of complex system science, machine learning and biology seem to suggest a list of necessary "ingredients" that such a system should comply with to have the hope of exhibiting a high level of open-ended dynamics [Stanley et al., 2020]. The next section details what we include as *desirata* properties in our open-ended framework.

# Desirata Properties of our Open-Ended Framework

## $G$: self-organized system for the endless generation of complex artifacts

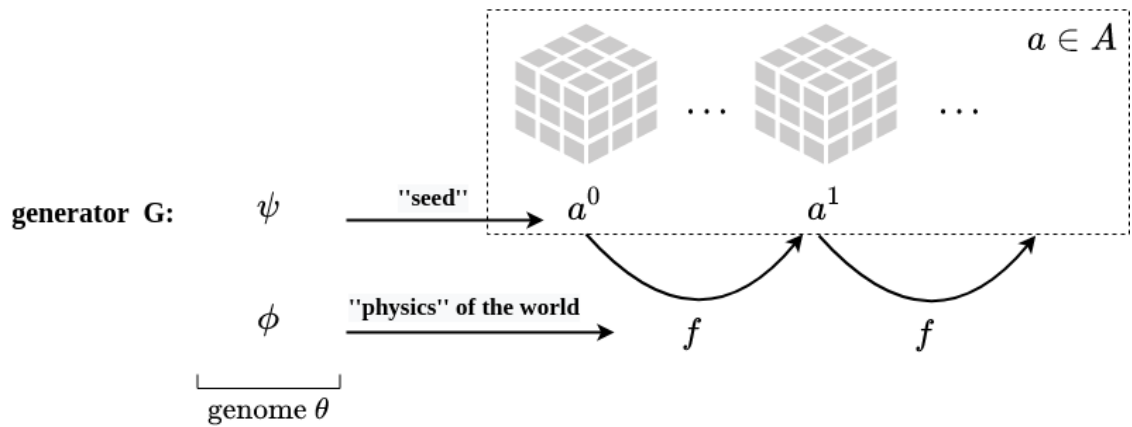

We formulate our *artifact-generator $G$* as an artificial complex system which should comply with the following general properties:

- **Dynamical system** — $G$ recursively applies an update rule $f$ that changes the state of the system over time.
- **Growth** — the initial state $a^0$ is limited in space and complexity and the phenotype grows and complexifies over time.
- **Local interactions** — the update rule $f$ is defined on a local neighborhood ball and shared at the different locations.
- **Non-linearity** — $G$ has non-linear dynamics $f$.
- (may be) **Open** — G might exchange flows of matter/energy with the environment to allow high levels of diversity, heterogeneity and complexity.
- (may have a) **Memory** — G might memorize the system's past states into the *hidden* features to allow for more advanced behaviors.

The idea of implementing computational systems with the above *desirata* properties, whose complexity could grow automatically akin to real complex systems is not new. The simulation of artificial worlds that exhibit life-like behaviors goes back to von Neumann's seminal work on self-reproducing cellular automata in the 40's. and has been one of the core thematic of the Artificial Life research discipline since then.
Interestingly, von Neumann's "proof of principle" showed that very simple machines could constitute universal constructors with very large expressive power. 30 years later, John Conway implemented a very simple version of von Neumann's automata, the well known Game of Life that, despite its apparent simplicity, can generate a very wide range of life-like structures and dynamics.

Yet, even for very simple systems like Conway's Game of Life where one fully knows the basic rules at the local level, we are still far from fully grasping what structures can self-organize, how to represent and classify them, and how to predict their evolution. As an illustration of those challenges, more than 50 years after the creation of Conway's game of like and thanks to today supercomputer's power, the community is still discovering novel patterns with impressive complexity and whose existence remained unknown to date.

Therefore, a more challenging question is:

> *Can we automate the long-term discovery of increasingly complex and divergent structures in such systems?*

In the same way that Conway's Game of Life together with the community of people searching it might form an example of open-ended system, our open-ended framework couples the artifact-generator $G$ with an artifact-discovery algorithm $D$ able to continuously search for novel and divergent artifacts in $W$.

In the next section, we briefly present the 3 main families of methods that propose to *automatically* explore the input space $\Theta$ of complex systems, namely *"naive"* strategies (independent of the outcomes of the system), *optimization-driven* strategies and *diversity-driven* strategies.

We review why these approaches (in their standard definition) are limited to comply with open-ended problems and rather tend to converge toward a sub-region of the possible solution space $W$. Our *artifact-discovery* algorithm $D$ search objective is formulated as the *meta-diversity* objective, an extension of the standard *diversity* objective that we proposed recently in the context of automated discovery in complex systems [Etcheverry et al. 2020].

## $D$: meta-diversity search for the endless discovery of novel artifacts

### Naive search strategy

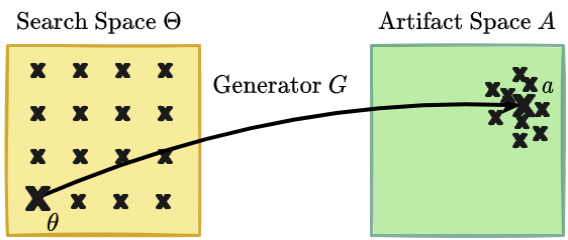

An "naive" strategy consists of covering at best the known search space $\Theta$, for instance with *grid-search* or *uniform* sampling strategy $D : \theta \sim U(\Theta)$. While easy to implement, this is often very inefficient in covering the space of possible phenotypes, especially for complex genotype-to-phenotype developmental mappings where big regions of the input space $\Theta$ are mapped to small regions in the outcome space $A$ (we talk of "attractor" states).

## Optimization search strategy

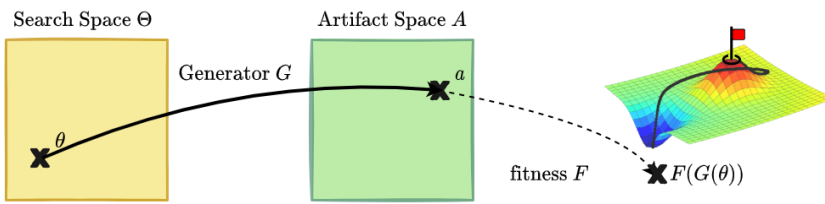

Pure objective-driven search is another widely-used discovery strategy $D$, where each discovery $a$ gets scored with a fitness function $F$ and where $D$ aims to find individuals $\{(\theta_i, a_i)\}$ with the highest fitness scores $F(a_i)$. Different objective-driven techniques have been implemented as discovery tool in complex systems such as evolutionary population-based search strategies [Nichele 2016], gradient-based search strategies (if the generator $G$ and the fitness $F$ are differentiable) [Mordvintsev et al. 2020], and reinforcement learning strategies (if the system's dynamics $f$ integrates a differentiable policy which actions influence the system next state) [Pathak et al. 2019]. While these techniques can be very powerful optimizers toward desired artifacts [Sudhakaran et al. 2021], they tend to converge toward "peaks" regions on the artifact fitness landscape and stop here once reached, making them poor-fit for open-ended problems. Moreover, they generally suffer from deceptive rewards and are hardly generalizable to complex artifact spaces with limited predictability.

## Diversity search strategy

Another family of approaches proposes to maximize the *novelty* of the discovered phenotypes instead of the *fitness*, and leverages population-based evolutionary and developmental curiosity-driven exploration methods like Novelty Search or

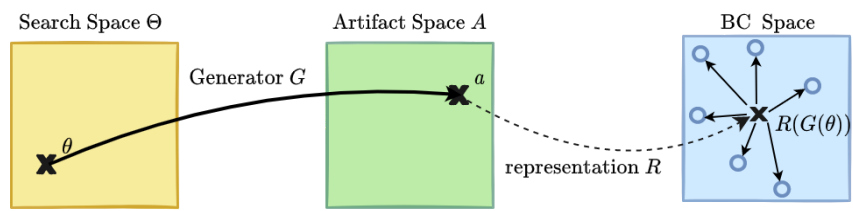

**Intrinsically Motivated Goal Exploration Processes.** Getting rid of the efficiency objective has shown to be effective at creating behavioral *diversity* in robotic systems [Forestier et al., 2017] and a promising discovery tool in complex systems [Reinke et al. 2020]. Moreover, diversity-search can be coupled with optimization-based strategies to generate a set of diverse high-performing phenotype solutions to a problem, forming what we call Quality-Diversity methods. Yet standard approaches assume that the intuitive notion of diversity can be captured within a single representation space, which is generally called $BC$ for behavioral characterization. This limits the scope of the final discoveries: in the same way that the fitness function $f$ is constraining the reached area in the observation space $A$, being diverse in a specific $BC$ does not mean being diverse in $A$. Actually, there is not such thing as a unique ground truth "interesting" diversity in complex phenotypic spaces, and operating in a monolithic BC space might lead to discoveries that are highly-diverse in that space but poorly-diverse according to other potentially-interesting behavioral criteria [Etcheverry et al. 2020].

## Meta-Diversity search strategy

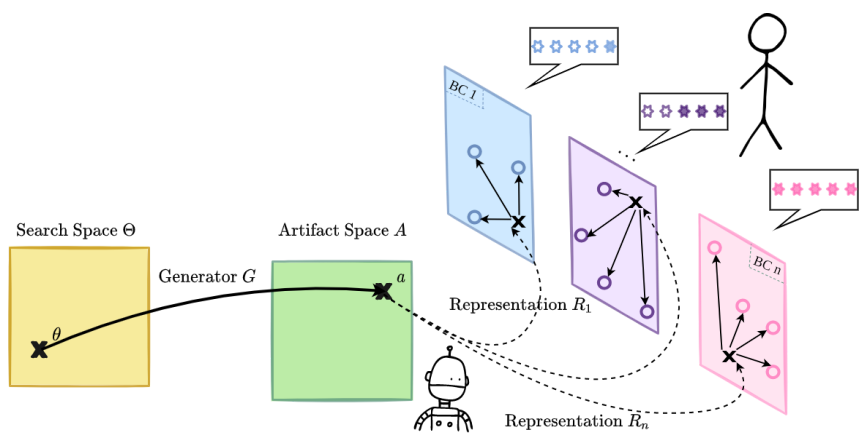

To address these limits, we recently formulated the problem of meta-diversity search where an artificial "discovery assistant" incrementally learns divergent feature spaces to characterize the different niches of diversities (outer loop) and searches to discover diverse patterns within each of them (inner loop). The objective of this process is to enable continuous seeking of novel source of diversities while being able to quickly adapt the search toward a new unknown type of diversity. Similarly, *quality-guidance* can be integrated in the framework: with minimal external feedback, a successful discovery assistant should be able to efficiently specialize the exploration strategy toward a particular type of "interesting" diversity, corresponding to the initially unknown preferences of an end-user and expressed through simple *sparse* feedback.

## Proposed Implementation

We simulate an artificial "chemistry" system based on Lenia continuous cellular automaton [Chan 2019, Chan 2020] and an artificial "discovery assistant" (HOLMES) which integrates a goal-based intrinsically-motivated exploration (*inner loop*'s diversity search) with the incremental learning of a growing hierarchy of behavioral characterization spaces (*outer loop*'s divergent knowledge accumulator) [Etcheverry et al. 2020].

While abstracting away from natural chemistry, we follow this metaphor to illustrate our open-ended system and implementation choices for the challenge.

### 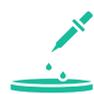 LeniaChem: an "artificial chemistry" system

Cellular automata are good models to study how significantly different dynamics can arise from uniform simple laws at a smaller scale.

Lenia is a recently proposed cellular automaton that extends traditional CA with continuous state spaces and rules generalized to higher dimensions, multiple kernels and multiple channels. Our variant of Lenia (LeniaChem) used for the Minecraft environment is implemented in 3D and combines continuous channels (one channel per "block" in minecraft) and discrete visible states. Any number kernels can be sampled and the kernel matrices are generated with CPPN networks [Stanley, 2007], such that the CA update rule is not limited in its expressive power.

The CA's 3D grid $a^t$ represents the "petri-dish" where evolved patterns will develop in time.

The state space $\Sigma$ is represented with a multi-channel continuous array $A$ where each channel $A_j$ represent the "chemical potential" of the associated specie

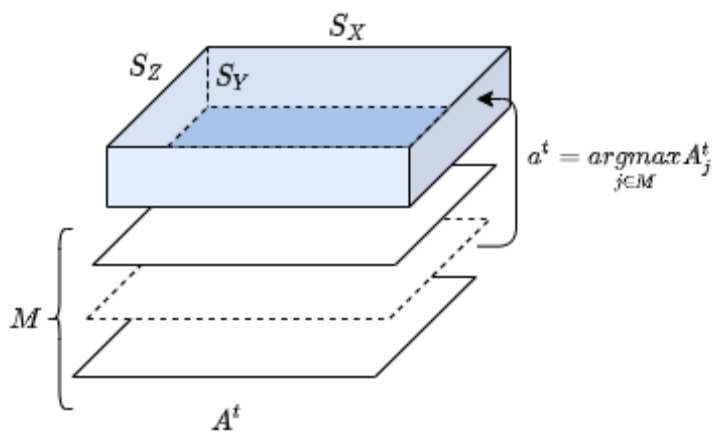

The hyper-parameters constraining the range of possible patterns are the grid-size $(S_X, S_Y, S_Z)$ (dimensions of the "petri-dish") and the number of channels $M$ (number of "chemical species" represented with blocks in Minecraft). The grid here is a torus, meaning that voxels on the top borders are neighbors to bottom voxels and same for left and right borders.

**Parameters $\psi$ that determine $A^{t=0}$:**

- a binary vector $I$ of shape $M$ which determines which channel get initialized (i.e. non-empty mass at t=0)
- Each initially-present channel $j$ ($I_j = 1$) is described with:
  - its *occupation ratio*: portion of the grid where it is present at t=0
  - its *cppn genome* which is used to generate the initial pattern $A_j^{t=0}$

**Parameters $\phi$ of the CA's update rule $f$:**

- a binary vector $C$ of shape $M^2$ which determines which species are interacting together
- Each interacting channel pair $(c_i, c_j)$ ($C_{i \to j} = 1$) is described with:
  - fraction $h$ of the update applied per time step (intra-specy rate of growth)
  - the kernel $K$ with:
    - the kernel size $R_x, R_y, R_z$ (*neighborhood* size)
    - the *cppn genome* which generates the kernel matrix $K$
  - the *growth function* mean $\mu$ and variance $\sigma$
- fraction $dt$ of the growth update applied per time step (global rate of growth)

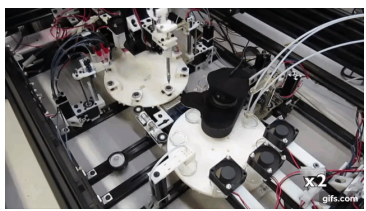

Following the "artificial chemistry" metaphor, we can imagine the parameters $\theta = (\psi, \phi)$ as representing the quantity and disposition of each starting material as well as the operating conditions (temperature, pressure, pH, etc.) influencing the development of the final product. As an example, Grizou et al. 2020 recently propose to explore the dynamics of an oil-droplet system with a "curious robot" which could manipulate the initial mixture composition of chemicals.

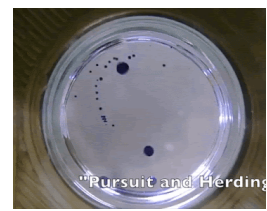

The resulting "artifact" here would be the yield of each compound in the final product, the rest being considered as empty "air" invisible blocks in Minecraft.

The **"physics" formulas** $f$ of our petri-dish follow Lenia rules [Chan 2020] and apply the following updates to the state at each time step:

1. Loop on all kernel $K_{i \to j}$ and:
   1. calculate the weighted sums $K_{i \to j} * A_i$ with its source channel $A_i$
   2. Apply growth mapping to the weighted sums:
   $$G_{i \to j} = \sum 2 \times e^{-\frac{\left((K_{i \to j}*A_i)-\mu\right)^2}{2\sigma^2}} - 1$$
   3. Add a small relative portion $dt * h_{i \to j}$ of the resulting growth to the destination channel: $A_j \mathrel{+}= dt * h_{i \to j} * G_{i \to j}$

2. The potentials of each channels $A_j$ are clipped between 0 and 1

3. To prevent participation of empty "air" cells in the update process we apply an alive-masking step [Mordvintsev et al. 2020] as follows: we fix the air potential (channel 0) to a fixed value 0.1 and we consider empty every location which maximum potential is lower than 0.1, by setting every channels (except air's one) to 0.

4. Finally chemical "matter" is created in zones of higher chemical potential with $a^t = \underset{j \in M}{argmax} A^t$

Back to the *desirata* properties of our artifact-generator, LeniaChem is a *dynamical* system that shows *growth*, *local interactions* and *non-linearity*. LeniaChem is not *open* in a strictly-speaking sense but we do not impose any sort of mass-conservation law. Similarly, we do not explicitly implement *memory* in the CA but any value can be stored in the continuous channel.

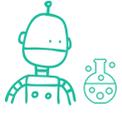

**Holmes: an artificial "discovery assistant" exploring the system**

To solve the previously-introduced meta-diversity objective, several key challenges need to be addressed. A first challenge is to unsupervisedly learn a *diverse* set of representations to characterize behaviors and a second challenge is to find a diverse set of patterns in each of those BC spaces.

## 1) Outer-loop: incremental learning of divergent characterization spaces

To address the first challenge, we use a dynamic neural-network architecture where a hierarchy of module embedding networks $\mathcal{R} = \{\mathcal{R}_i\}$ is progressively expanded by the discovering agent to characterize the different discovered niches of pattern. The architecture has 4 main components:

(i) a base *module* embedding neural network
(ii) a *saturation* signal that triggers the instantiating of new nodes in the hierarchy
(iii) a *boundary* criteria that unsupervisedly clusters the incoming patterns into the different modules
(iv) a *connection-scheme* that allows feature-wise transfer from a parent module to its children

In HOLMES, the clustering of the different patterns and the use of learnable connections is central to explore **divergent search spaces** and to **aggregate the knowledge** between the distributed modules. While HOLMES architecture is very general, the engineer's choice for the module/connections training strategy and for the clustering algorithm will impact how patterns are separated and distributed into different niches, which will in turn greatly influence their "selectivity" by the population-based IMGEP.

Here, we use the same implementation choices that in Etcheverry et al. 2020:
(i) we use VAEs as base modules $\mathcal{R}_i$: the VAEs are incrementally trained to encode their own niche of patterns into a latent characterization space $BC_i$
(ii) we say that the representational capacity of a module *saturates* when the reconstruction loss of its VAE reaches a plateau (with additional conditions to prevent premature splitting such as minimal node population)
(iii) each time a node gets saturated, we freeze its latent space and use K-Means

clustering to fit a boundary and redirect the incoming patterns (the boundary is kept fixed for the rest of the exploration)

(iv) we instantiate learnable layers called "lateral connections" between the parent and child VAE feature-maps allowing the child VAE to reuse its parent knowledge while learning to characterize novel dissimilar features in its own BC space

## 2) Inner-loop: find a diverse set of patterns in each characterization space

A goal-based intrinsically-motivated exploration process (IMGEP) is used for the parameter sampling strategy. The IMGEP operates in the hierarchy of goal spaces $BC_i$ as defined by HOLMES embedding hierarchy $\mathcal{R}_i$.

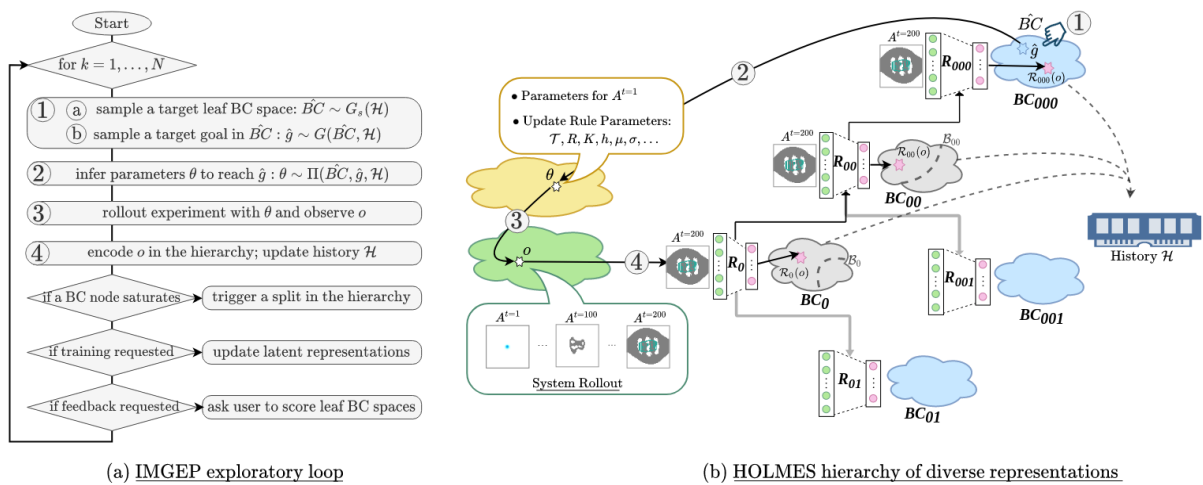

(a) IMGEP exploratory loop  (b) HOLMES hierarchy of diverse representations

The exploration process iterates through steps 1-to-4 as illustrated in the above figure:

(i) sample a target BC space and a target goal in it (the sampling strategy here is equivalent to Novelty Search)

(ii) sample a set of parameters for the next rollout to achieve that goal (here nearest neighbor + mutation)

(iii) let the system rollout and observe the outcome artifact

(iv) store the resulting (parameter, observation, encodings) triplets in an explicit memory (growing archive of all discoveries).

## Discussion

In the context of such an "open-ended challenge" were the ambitions were also quite "open-ended", we were more or less free to do anything we want. It has been quite fun implementing our ideas in the Minecraft environment, but there are many other directions that we could explore for future work.

While we mainly used the Minecraft engine as a "renderer" in our current experiments, it will be interesting that our growth process occurs inside the world's physics. Minecraft physics are quite poor (most blocks don't even have gravity) but some blocks do have interesting properties and we could also imagine having external sources perturbing the growth process (eg: players throwing blocks inside the petri-dish).

Secondly, it will be interesting to integrate some sort of quality-guidance in the meta-diversity search, for instance by rewarding displacements of the structure's center of mass to hope discovering flying machines in Minecraft [Grbic et al., 2020], but defining such score might be hard for more complex functionalities (how to reward building a CPU machine?) and might be very difficult to optimize due the sparse reward.

Another type of guidance that would be very interesting to integrate in the discovery process is human-guidance. We have shown in Etcheverry et al. 2020 that minimal feedback from an external user could help the meta-diversity search specialize toward "interesting" types of diversity. What constructions would come out from a community of players guiding the discoveries in distributed parts of the world with their subjective notion of what is "interesting"?

## Acknowledgements

This work benefitted from access to the HPC resources of IDRIS under the allocation 2020-[A0091011996] made by GENCI, using the Jean Zay supercomputer.

## Resources

The source code used for this challenge can be found at the following link: https://github.com/mayalenE/evocraftsearch/